\documentclass[11pt,a4paper]{article}
\usepackage[hyperref]{acl2019}
\usepackage{times}
\usepackage{latexsym}

\usepackage{amsmath}

\usepackage{url}
\usepackage{graphicx}
\usepackage[utf8]{inputenc}

\aclfinalcopy

\title{Ask to Learn: A Study on \emph{Curiosity}-driven Question Generation}
\author{Thomas Scialom\textsuperscript{\rm *‡}, Jacopo Staiano\textsuperscript{\rm ‡}\\
\textsuperscript{\rm *} Sorbonne Université, CNRS, LIP6, F-75005 Paris, France\\
\textsuperscript{\rm ‡} reciTAL, Paris, France\\
\texttt{\{thomas,jacopo\}@recital.ai}
}
 
\begin{document}

\maketitle

\begin{abstract}
We propose a novel text generation task, namely \emph{Curiosity}-driven Question Generation. We start from the observation that the Question Generation task has traditionally been considered as the dual problem of Question Answering, hence tackling the problem of generating a question given the text that contains its answer. Such questions can be used to evaluate machine reading comprehension. However, in real life, and especially in conversational settings, humans tend to ask questions 
with the goal of \emph{enriching} their knowledge and/or \emph{clarifying} aspects of previously gathered information. We refer to these inquisitive questions as \emph{Curiosity}-driven: these questions are generated with the goal of obtaining new information (the answer) which is not present in the input text. In this work, we experiment on this new task using a conversational Question Answering (QA) dataset; further, since the majority of QA dataset are not built in a conversational manner, we describe a methodology to derive data for this novel task from non-conversational QA data. We investigate several automated metrics to measure the different properties of \emph{Curious Questions}, and experiment different approaches on the \emph{Curiosity}-driven Question Generation task, including model pre-training and reinforcement learning. Finally, we report a qualitative evaluation of the generated outputs. 

\end{abstract}

\section{Introduction}
The growing interest in Machine Reading Comprehension (MRC) has sparked significant research efforts on Question Generation (QG), the dual task to Question Answering (QA). In QA, the objective is to produce an adequate response given a query and a text; conversely, for QG, the task is generally defined as generating relevant question given a source text, focusing on a specific answer span. 
To our knowledge, all works tackling QG have thus far focused exclusively on generating relevant questions which can be answered given the source text: for instance, given \emph{AAAI was founded in 1979} as input, a question likely to be automatically generated would be \emph{When was AAAI founded?}, where the answer \emph{1979} is a span of the input. Such questions are useful to evaluate reading comprehension for both machines \cite{hermann_teaching_2015,eyal_question_2019-1} and humans \cite{mani_tipster_1999}.

However, the human ability of asking questions goes well beyond evaluation: asking questions is essential in education \cite{gall_use_1970} and has been proven to be fundamental for children cognitive development \cite{chouinard_childrens_2007}. Curiosity is baked into the human experience. It allows to extend one's comprehension and knowledge by asking questions that, while being relevant to context, are not directly answerable by it, thus being \emph{inquisitive} and \emph{curious}. The significance of such kind of questions is two-fold: first, they allow for gathering novel relevant information, \emph{e.g.} a student asking for clarification; second, they are also tightly linked to one's understanding of the context, \emph{e.g.} a teacher testing a student's knowledge by asking questions whose answers require a deeper understanding of the context and more complex reasoning.

From an applicative point of view, we deem the ability to generate curious, inquisitive, questions as highly beneficial for a broad range of scenarios: i) in the context of human-machine interaction (e.g. robots, chat-bots, educational tools), where the communication with the users could be more natural; ii) during the learning process itself, which could be partially driven in a self-supervised manner, reminiscent of how humans learn by exploring and interacting with their environment.

To our knowledge, this is the first paper attempting to tackle \emph{Curiosity}-driven neural question generation. 
The contributions of this paper can be summarized as follow:
\begin{itemize}
    \item we propose a new natural language generation task: \emph{curiosity}-driven question generation;
    \item we propose a method to derive data for the task from popular \emph{non-conversational} QA datasets;
    \item we experiment using language model pre-training and reinforcement learning, on two different datasets;
    \item we report a human evaluation analysis to assess both the pertinence of the automatic metrics used and the efficacy of the proposed dataset-creation method above.
\end{itemize}

\section{Related Works}
Deep learning models have been widely applied to text generation tasks such as machine translation~\cite{kalchbrenner_recurrent_2013}, abstractive summarization~\cite{rush_neural_2015} or dialog~\cite{henderson_deep_2013}, providing significant gains in performance.
The state of the art approaches are based on sequence to sequence models \cite{cho_learning_2014,sutskever_sequence_2014}.
In recent years, significant research efforts have been directed to the tasks of Machine Reading Comprehension (MRC) and Question Answering (QA) \cite{hermann_teaching_2015,rajpurkar_squad:_2016}. The data used for tackling these tasks are usually composed of $\{context, question, answer\}$ triplets: given a context and the question, a model is trained to predict the answer.

Conversely, the Question Generation (QG) task introduced by \cite{du_learning_2017,zhou_neural_2017} can be considered as the dual task for QA \cite{duan2017question}: thus, given a context and (optionally) an answer, the model is trained to generate the question. Following QA, research on QG \cite{amidei_evaluation_2018} has also seen increasing interest from the community. One of the main motivations is that an effective QG model can be used to generate synthetic data in order to augment existing QA datasets \cite{yuan_machine_2017,alberti_synthetic_2019}. 
For instance, \cite{yuan_machine_2017} proposed a reinforcement learning setup trained using a QA-based metric reward: given a paragraph and an answer, the model first generates questions; then, the paragraph and the corresponding generated questions are given to a pre-trained QA model which predicts an answer; finally, the reward is computed as the number of overlapping words between the ground truth answer and the predicted answer. 
For an extensive evalution of models trained with different rewards we refer the reader to \cite{hosking_evaluating_2019}. Most of  these works followed \cite{ranzato_sequence_2015}, who applied reinforcement to neural machine translation. First, a sequence to sequence model is trained under teacher forcing \cite{williams_learning_1989} to optimize cross-entropy, hence helping to reduce the action space (i.e. the vocabulary size). Then, the model is finetuned with a mix of teacher forcing and REINFORCE \cite{williams_simple_1992}.

For automatic evaluation, all previous works on QG resort to BLEU metrics \cite{papineni_bleu:_2002}, originally developed and widely used in Machine Translation. However, how to evaluate text generation models remains an open research question:  \cite{nema_towards_2018} pointed out that, on QG tasks, the correlation between BLEU and human evaluation was poor. 

A thorough investigation of the behavior of open-domain conversational agents
has been recently presented by \cite{see_What_2019}. Using controllable neural text generation methods, the authors control important attributes for chit-chat dialogues, including question-asking behavior. Among the take-away messages of this work, is that question-asking represents an essential component in an engaging chit-chat pipeline: the authors find, via a large-scale human validation study, that agents with higher rates of question-asking obtain qualitative improvements in terms of inquisitiveness, interestingness and engagingness.

Indeed, in a conversational setting, it can be expected that the nature of follow-up questions significantly differs from those used as target in a traditional QG training setup: as mentioned earlier, QG has so far been tackled as the dual task to QA, hence training models to generate questions whose answer is present in the input context. On the contrary, we argue that in natural conversations the questions \emph{follow} the input context but are rather a mean to augment one's knowledge (thus, their answer is \emph{not} present in the input context). In this work, we thus define the task as \emph{Curiosity}-driven Question Generation.

\section{Dataset}
\label{sec:Dataset}

Question Answering datasets are usually composed of a set of questions  associated with the corresponding answers and the reading passages (the \emph{context}) containing the answer. The QA task is defined as finding the answer to a question given the context. As opposed, the Question Generation (QG) task is to generate the question given the input and (optionally) the answer. 
Most previous efforts on the QG task have resorted to the widely used Stanford Question Answering Dataset (SQuAD)~\cite{rajpurkar_squad:_2016}. It contains roughly 100,000 questions posed by crowd-workers on selected sample of Wikipedia articles. Several other QA datasets have also been recently published accounting for characteristic such as requiring multi-passage or discrete reasoning \cite{yang2018hotpotqa,Dua2019DROP}; further, \emph{conversational} QA datasets have been made available: CoQA \cite{reddy2019coqa} and QuAC \cite{choi-etal-2018-quac} have the desirable property to be in a dialogue-like setting. 

In our scenario, \emph{Curiosity}-driven QG, the reading passage associated with a question should \emph{not} contain the answer, but rather pave the way for asking a new question -- whose answer would eventually enrich the knowledge on the matter at hand. Therefore, a natural choice to build QG data would be to rely on existing datasets for \emph{conversational} QA. A detailed comparison of the above-mentioned CoQA and QuAC datasets is provided by \cite{yatskar_qualitative_2019}, who reports the proportion of \emph{Topic Error} (questions unlikely to be asked in the context) and \emph{Entity Salad} (i.e. questions unanswerable for any context\footnote{see section 2.1 in \cite{yatskar_qualitative_2019}}): CoQA includes a significantly higher proportion \emph{Topic Error} and \emph{Entity Salad} compared to QuAC. For this reason, we resort to QuAC in order to derive data \emph{Curiosity}-driven QG.

Furthermore, recognizing the fact that the great majority of QA datasets available does not account for conversational characteristics, we propose a methodology to derive data for \emph{Curiosity}-driven Question Generation from standard QA datasets, applying it to the popular SQuAD \cite{rajpurkar_squad:_2016}.

For both our data sources, and consistently with standard QA and QG tasks, we encode each sample as a triplet $\{P, q, a\}$ where the paragraph $P$ comprises $n$ sentences $[s_0,..., s_n]$, and $a$ represents the answer to the question $q$. 
A canonical QG approach would thus use $s_a$, i.e. the sentence of $P$ that contains the answer, as source, and $q$ as generation target.
On the contrary, for \emph{Curiosity}-driven QG, any sentence $s_x$ from $P$ can potentially be used as the source sequence, as long as it does not contain the answer -- i.e. under the necessary constraint of $x \neq a$. In the following subsections, we elaborate on additional constraints depending on the nature of the source data.

In general, we define samples as triplets
\begin{equation}
\label{eq:triplet}
t = \{s_x, P', y\}
\end{equation} 
where $s_x$ and $P'$ are, respectively, the input sentence and the paragraph $P$ modified according to the appropriate dataset-depending constraint, and $y$ is the reference (target) question.

\subsection{Conversational QA Data}
As mentioned above, we first derive our data from the QuAC dataset, which is built from Wikipedia articles by iterating over the following procedure: given a sentence, a student annotator asks a relevant question for which he does not have the answer; then, the teacher -- annotator -- retrieves a sentence that contains the answer. Thus, a QuAC question is \emph{curious} by design, given the text that precedes it. More formally, for the question $q$ (i.e. our target), the source $s_x$ is composed by the concatenation of the sentences of $P$ which appear before the sentence $s_a$ that contains the answer. Therefore, our QuAC-derived dataset is built by applying the stricter constraint $x < a$.

Numerically, the QuAC dataset compounds to 83,568 questions (on 11,567 articles) for the train set, 7,354 for the validation set and 7,353 for the test set (1,000 articles each). Since the test set is not public, we use the original QuAC validation set to build our test set. From the training set, we randomly drop 1,000 articles (hence, 7,224 samples) which we use to derive our validation set, thus resulting in 76,345 questions for training. 

\subsection{Standard QA Data}
Most of the available QA datasets are not conversational. Thus, we propose a simple method to obtain data for \emph{Curiosity}-driven QG from standard QA datasets. For this, we use the widely popular SQuAD\cite{rajpurkar_squad:_2016}, and specifically the original splits released by \cite{du_learning_2017} which is commonly used for Question Generation.

As opposed to QuAC, the questions in SQuAD do not follow logical ordering. Therefore, any sentence $s_x$ from $P$ can potentially be used as the source sequence, as long as it does not contain the answer $a$ (constraint: $x \neq a$). 
Nonetheless, as is reasonable for factoid QA datasets, several questions are so specific to their associated sentence $s_a$ that they would be extremely unlikely to be asked without knowing the contents of $s_a$ itself.

To exemplify this issue, take the following paragraph from SQuAD:
\begin{center}
\noindent\fbox{%
    \parbox{.96\columnwidth}{%
        Tesla was the fourth of five children. He had an older brother named Dane and three sisters, Milka, Angelina and Marica. Dane was killed in a horse-riding accident when Nikola was five. In 1861, Tesla attended the ``Lower" or ``Primary" School in Smiljan where he studied German, arithmetic, and religion. In 1862, the Tesla family moved to Gospić, Austrian Empire, where Tesla's father worked as a pastor. Nikola completed ``Lower" or ``Primary" School, followed by the ``Lower Real Gymnasium" or ``Normal School.
        
    }%
}
\end{center}

Given \emph{``Dane was killed in a horse-riding accident when Nikola was five."} as $s_a$, and operating under the sole constraint of $x \neq a$, the sentence \emph{``Tesla was the fourth of five children"} would be eligible as a source $s_x$ for the target question \emph{``What happened to Dane?"}. This question can only be asked if either contextual information or background knowledge is available, since it requires to know that \emph{Dane} was among Tesla's four siblings.

To overcome this problem, we added an additional constraint based on Named Entity Recognition (NER): 
$s_x$ is an acceptable input only if all the entities present in the question $q$ are also present in the input sentence $s_x$. In the previous example, this would thus filter out the target \emph{``What happened to Dane?"} while allowing for \emph{``What was Tesla's brother's name?"}.

For our experiments we used \texttt{spaCy}\footnote{\url{https://spacy.io/usage/linguistic-features}}.

\begin{table}
\centering
\begin{tabular}{|l|c|c|c|}
\hline
                                 & \textbf{Train} & \textbf{Dev} & \textbf{Test} \\ \hline \hline
\textbf{Learning to ask}            & 86,635         & 8,965        & 8,964         \\ \hline \hline
\textbf{Unconstrained} & 342,768        & 27,624       & 27,807        \\ \hline
\textbf{Constrained}     & 25,356         & 2,076        & 2,087         \\ \hline
\end{tabular}
\caption{Data distributions over the train-validation-test splits. \emph{Learning to ask} refers to the original split released by \cite{du_learning_2017}, from which our data is derived. The bottom rows refer to the data we obtain using our methodology, with and without NER constraining.}
\label{tab:dataset}
\end{table}

In Table~\ref{tab:dataset} we report the number of samples we obtained from SQuAD before and after applying NER filtering. After applying the above methodology to construct a dataset for \emph{Curiosity}-driven QG, our training dataset contains 25,356 samples for training, 2,076 for development, and 2,087 for testing.

\section{Metrics}

Automatic evaluation of Natural Language Generation (NLG) systems is a challenging task \cite{nema_towards_2018}. For QG, $n$-gram based similarity metrics are commonly used. These measures evaluate how similar the generated text is to the corresponding reference(s). While they are known to suffer from several shortcomings \cite{paulus_deep_2017,liu_how_2016}, they allow to evaluate specific properties of the developed models. In this work, the metrics detailed below are proposed and we evaluate their quality through a human evaluation in subsection \ref{subsec:human_evaluation}.

\subsection{BLEU}
One of the most popular metrics for QG, BLEU \cite{papineni_bleu:_2002} provides a set of measures to compare automatically generated texts against one or more references. In particular, BLEU-N is based on the count of overlapping \emph{n}-grams between the candidate and its corresponding reference(s). 

\subsection{Self-BLEU}
Within the field of Computational Creativity, \emph{Diversity} is considered a desirable property \cite{karampiperis_towards_2014}. Indeed, generating always the same question such as \emph{``What is the meaning of the universe?"} would be an undesirable behavior, reminiscent of the ``collapse mode" observed in Generative Adversarial Networks (GAN) \cite{Goodfellow_generative_2014}. 
Therefore, we adopt Self-BLEU, originally proposed by \cite{zhu_texygen:_2018}, as a measure of diversity for the generated text sequences. Self-BLEU is computed as follows: for each generated sentence $s_i$, a BLEU score is computed using $s_i$ as hypothesis while the other generated sentences are used as reference. When averaged over all the references, it thus provides a measure of how diverse the sentences are. Lower Self-BLEU scores indicate more diversity. We refer to these metrics as \textbf{Self-B*} throughout this paper.

\subsection{QA-based metrics}

Given a text, a question can be considered \emph{curious} if the answer is not contained in the input text. In our task, this implies that a question $q$ should not be answerable given its corresponding input sentence $s_x$. Thanks to the recent improvements obtained on Question Answering tasks -- for instance, human-level performance has been achieved on SQuAD-v1\footnote{\url{https://rajpurkar.github.io/SQuAD-explorer/}} -- the \emph{answerability} of a question can be automatically measured.

Therefore, given a question-context pair as input to a QA model, two type of metrics can be computed: 
\begin{enumerate}
    \item{\emph{n-gram based score}: measuring the average overlap between the retrieved answer and the ground truth.}
    \item{\emph{probability score}: the confidence of the QA model for its retrieved answer; this corresponds to the probability of being the correct answer assigned by the QA model to the retrieved answer.}
\end{enumerate}

Since several diverse questions can be generated for a given input, we consider the latter metric (\emph{probability score}) to better fit the \emph{Curiosity}-driven QG task. 

Hence, given the evaluated question $q$ and the input text $s_x$, we define a metric \emph{QA\_{prob}} as the confidence of the QA model that its predicted answer is correct. This metric measures \emph{answerability} of $q$ given $s_x$: therefore, the lower this score, the less likely the answer is contained in the input text. 

While being non-answerable represents a necessary condition for $q$ being a \emph{curious} question with respect to its context $s_x$, we also want $q$ to be as relevant and useful as possible. To this end, we compute the above \emph{QA\_{prob}} for question $q$ on $P'$, which represents the source paragraph stripped from the sentence containing the answer (see Eq. \ref{eq:triplet}). The higher this score, the more likely the question is relevant and useful to augment the knowledge provided by $s_x$.

Thus, the two proposed metrics are defined as
\begin{equation}
    QA_{source} = QA_{prob}(q, s_x)
\end{equation}
and
\begin{equation}
    QA_{context} = QA_{prob}(q, P')
\end{equation}

Under our definition, \emph{Curiosity}-driven questions are those that minimize $QA_{source}$ while maximizing $QA_{context}$.
To compute these QA-based metrics, we use the HuggingFace implementation\footnote{\url{https://github.com/huggingface/pytorch-transformers}} of BERT \cite{devlin_bert:_2018}.

\section{Experiments}

\subsection{Baseline model}
As baseline architecture we adopt the popular Transformer \cite{vaswani_attention_2017}, which proved to perform well on a wide range of text generation tasks. Among these, neural machine translation \cite{ott-etal-2018-scaling}, automatic summarization \cite{gehrmann_bottom-up_2018}, and question generation \cite{dong_unified_2019,scialom-etal-2019-self}. It can be briefly described as a sequence-to-sequence model with a symmetric encoder and decoder based on a self-attention mechanism, which allows to overcome the inherent obstacles to parallelism present in recurrent models such as Long Short Time Memory (LSTM) networks \cite{hochreiter_long_1997}. 

The copy mechanism \cite{gulcehre_pointing_2016} proved beneficial for QG \cite{zhao_paragraph-level_2018,scialom-etal-2019-self}: indeed, the QG task is very sensitive to rare and out of vocabulary words such as named entities and such a mechanism help deal with it efficiently: more than 50\% of the answers in the SQuAD dataset, for instance, correspond to named entities (see Table 2 in \cite{rajpurkar_squad:_2016}. Hence, following \cite{gehrmann_bottom-up_2018,scialom-etal-2019-self}, we include a copy mechanism in our Transformer architecture. 

For our experiments, we used the following hyper-parameters for the transformer: 
\texttt{N = 2} (number of blocks);
\texttt{d\_model = 256} (hidden state dimension);
\texttt{d\_ff = 512} (position-wise feed-forward networks dimension);
and, \texttt{h = 2} (number of attention heads).

Experiments run with the original hyper-parameters\footnote{\texttt{N=6}, \texttt{d\_model=512}, \texttt{d\_ff=2048}, \texttt{h=8}.} as proposed by \cite{vaswani_attention_2017} obtained consistent and numerically similar results. During training, we used mini batches of size 64 and the Adam optimizer \cite{kingma_adam:_2014}. At generation time, the decoding steps are computed trough the beam search algorithm with $k=5$ beams by default.

\subsection{Reinforcement}
Reinforcement Learning (RL) is an efficient technique to maximize discrete metrics for text generation. Previously,
\cite{ranzato_sequence_2015} used the REINFORCE algorithm \cite{williams_simple_1992} to train RNNs for several generation tasks, showing improvements over previous supervised approaches. Moreover, \cite{paulus_deep_2017} combined supervised and reinforcement learning, demonstrating improvements over competing approaches both in terms of ROUGE and on human evaluation. 

However, the metrics used as reward are often overfit, leading to numerical improvements which do not translate to increased -- and, rather, contribute to degrading -- output quality, thus leading to reduced effectiveness of the trained models for practical applications. On this matter, and with a particular focus on QG, \cite{hosking_evaluating_2019} performed a human evaluation on RL models trained with several metrics as reward, finding them to be indeed poorly aligned with human judgments: the models appear to learn to exploit the weaknesses of the reward source.

To overcome this issue, we propose to use a balanced reward:
\begin{equation}
\label{eq:reward}
r(q, P, P') = QA_{context} - QA_{source}
\end{equation}
thus maximizing the probability of finding an answer to the generated question within the input paragraph but not inside the source sentence. 

In our experiments, we follow the approach proposed by \cite{ranzato_sequence_2015,paulus_deep_2017}, considering a mixed loss $L_{ml+rl}$ which combines supervised and reinforcement learning schemes:

\begin{equation}
\label{mlrl}
L_{ml+rl} = \gamma L_{rl} + (1 - \gamma) L_{ml}
\end{equation}

where the maximum likelihood $L_{ml}$ is defined as

\begin{equation}
L_{ml} = -\sum^{m}_{t=0}{log(p(y_t | y_0, ..., y_{t-1}, X))}
\end{equation}

where $X=[x_1,...,x_n]$ represents the source text of length $n$ and $Y=[y_1,...,y_m]$ the corresponding reference question of length $m$. 

Conversely, we define the reinforcement loss $L_{rl}$ to be \emph{minimized} according to the standard RL actor-critic scheme, where $r(q, P, P')$ is the reward function defined in \ref{eq:reward}:
\begin{equation}
L_{rl} =  (r(\widehat{Y}) - r(Y^s)) \sum^{m}_{t=0}{log(p(y^{s}_t | y^{s}_0, ..., y^{s}_{t-1}, X))}
\end{equation}

Greedy decoding according to the conditional distribution $p(y|X)$ is used to obtain a sequence $\widehat{Y}$. The model is sampled using its Markov property, that is, one token at a time, giving rise to the sequence $Y^s$.

\subsection{Pretraining (PT)}

As shown in Table~\ref{tab:dataset}, the constrained dataset amounts to roughly three times less samples than both QuAC and the original SQuAD dataset it derives from. We thus investigate, for this dataset, the effect of pretraining the model under the traditional (i.e. not \emph{Curiosity}-driven) QG training setup, using the training set as provided by \cite{du_learning_2017}). Then we resume training on the final dataset obtained after applying the NER-based constraint for \emph{Curiosity}-driven QG on the same training samples.

For the QuAC \emph{Curiosity}-driven dataset, the amount of data is comparable to the original dataset, given the \emph{conversational} nature of QuAC. Therefore, we do not use pretraining for the experiments on QuAC. 

\section{Results}

\subsection{Automatic metrics}

\begin{table*}
\begin{center}
\small
\begin{tabular}{l||c|c|c|c|c|c|c}
 & \textbf{human} & \textbf{base\_beam1} & \textbf{base\_beam3} & \textbf{base\_beam5} & \textbf{RL\_beam1} & \textbf{RL\_beam3} & \textbf{RL\_beam5}\\
 \hline \hline
\textbf{BLEU1} & - & 31.94 & 26.92 & 22.26 & 30.19 & 32.15 & 26.06\\
\textbf{BLEU2} & - & 14.45 & 14.76 & 13.55 & 13.19 & 16.01 & 15.28\\
\textbf{BLEU3} & - & 7.49 & 10.59 & 10.84 & 6.81 & 9.04 & 11.52\\
\textbf{BLEU4} & - & 4.31 & 8.79 & 9.59 & 3.72 & 6.1 & 9.85\\
\hline \hline
\textbf{Self-B1} & 96.09 & 99.84 & 99.88 & 99.95 & 99.96 & 99.94 & 99.96\\
\textbf{Self-B2} & 84.55 & 99.64 & 99.75 & 99.91 & 99.91 & 99.89 & 99.93\\
\textbf{Self-B3} & 70.55 & 99.39 & 99.63 & 99.87 & 99.86 & 99.84 & 99.9\\
\textbf{Self-B4} & 57.57 & 99.09 & 99.5 & 99.83 & 99.79 & 99.79 & 99.87\\
\hline \hline
\textbf{QA\textsubscript{source}} & 44.5 & 48.86 & 35.8 & 29.88 & 57.54 & 41.36 & 35.03\\
\textbf{QA\textsubscript{context}} & 48.94 & 48.32 & 40.96 & 38.48 & 55.38 & 42.95 & 41.63\\
\hline
\end{tabular}
\caption{Results obtained on QuAC-derived data.}
\label{tab:metric_results_quac}
\end{center}
\end{table*}

In Table~\ref{tab:metric_results_quac} we report the results of our experiments on QuAC for the baseline model (\emph{base}) and the RL model. We use a beam $k$, and compute the results for $k=[1,3,5]$.
In addition the generated questions with a beam $k=5$, we also computed the results for $k=1$ and $k=3$. While one would expect to see for all the metrics a slight improvement, with increasing beam size,
we observe a strong divergence among the results: 
increasing values for $k$ correspond to a significant improvements in terms of \emph{BLEU-4} 
and notable drops for \emph{BLEU-1}.
A similar phenomena was observed by \cite{ott_analyzing_2018} in the context of machine translation: in this work, the presence of 1 or 2\% of noisy data is found to be enough to significantly degrade the beam search results. In our case, one of most frequent generated question is \emph{Are there any other interesting aspects about this article ?}. Indeed, the frequency of this question in our training set amounts to 4.18\% of the questions. On the test set we see that roughly 80\% of the generated questions start with the token \emph{``are"} .
Generating this sequence is not very likely with a greedy search ($k=1$): at any time step during the generation, if any other token has a higher probability, this question will be dismissed. On the other hand, with a higher beam, it is likely to be kept and eventually result as the most probable sequence, among the different remaining beams at the end of the inference. 

\begin{table}
    \centering
    \small

\begin{tabular}{l|c||c|c|c|c}
                                & \textbf{human} & \textbf{base} & \textbf{RL} & \textbf{PT} & \textbf{PT+RL} \\ \hline \hline
\textbf{BLEU1}                  & -            & 32.81             & 31.71                & \textbf{33.02}            & 32.13      \\ \hline
\textbf{BLEU2}                  & -            & 14.31             & 13.67                & \textbf{14.9}            & 14.58      \\ \hline
\textbf{BLEU3}                  & -            & 7.57              & 7.21                 & \textbf{8.1}             & 7.81       \\ \hline
\textbf{BLEU4}                  & -            & 4.12              & 3.88                 & \textbf{4.61}             & 4.53       \\ \hline
\hline
\textbf{Self-B1}             & 95.85          & \textbf{93.80}    & 94.37                & 95.80            & 95.42               \\ \hline
\textbf{Self-B2}             & 87.96          & \textbf{87.00}    & 88.80                & 91.29            & 90.71               \\ \hline
\textbf{Self-B3}             & 81.75          & \textbf{79.59}    & 82.64                & 86.47            & 85.66               \\ \hline
\textbf{Self-B4}              & 77.60          & \textbf{72.60}    & 76.48                & 81.63            & 80.52               \\ \hline
\hline
\textbf{QA\textsubscript{source}}  & 54.12          & 57.85    & \textbf{55.87}       & 63.13            & 58.46               \\ \hline
\textbf{QA\textsubscript{context}} & 74.93          & 52.11             & \textbf{55.98}       & 50.81            & \textbf{56.36}      \\ \hline
\end{tabular}
\caption{Results obtained on SQuAD-derived data.}
\label{tab:metric_results_squad}
\end{table}

Moving to our SQuAD-based experiments, we observe that the models trained on SQuAD do not seem to suffer from this issue since all the metrics improved when increasing the beam size from $k=1$ to $k=5$. This is consistent with the results reported by \cite{zhao_paragraph-level_2018} where improving the beam improve slightly all the metrics. Thus, we only report the results with $k=5$ in Table~\ref{tab:metric_results_squad}. 
A possible explanation is that SQuAD, as opposed to QuAC, only contains factoid questions. 

We observe that the models trained with RL obtain, as could be expected, higher scores for \emph{QA\textsubscript{context}} with respect to those trained without RL. A higher \emph{QA\textsubscript{context}} implies that the QA model is more likely to find an answer in the near context of the source. \emph{QA\textsubscript{source}} is lower, as expected, for SQuAD based models, though comparatively higher than the models trained with RL on QuAC. We identify two possible reasons for this: first, the QA model is trained on answerable questions; second, the nature of the QUaC questions is less factoid than the SQuAD ones, and non-factoid questions can arguably be harder for the QA model to evaluate. This could explain why, in the RL setting, \emph{QA\textsubscript{context}} (the evaluation on answerable questions) is higher for both SQuAD and QUaC models, but only SQuAD models achieve a lower \emph{QA\_source} (the evaluation on non answerable questions). 

Furthermore, we see that pretraining allows to achieve higher BLEU scores, at the cost of lower Self-BLEU, thus showing an increased accuracy but less diversity in the generated questions. 
Indeed, we find that pretrained models tend to generate a higher number of questions starting with ``\emph{What}'' compared to both other models and the references; the distribution for the first words of the human questions appears closer to that non pretrained models. 

\begin{figure*}
    \centering
    \includegraphics[width=\columnwidth]{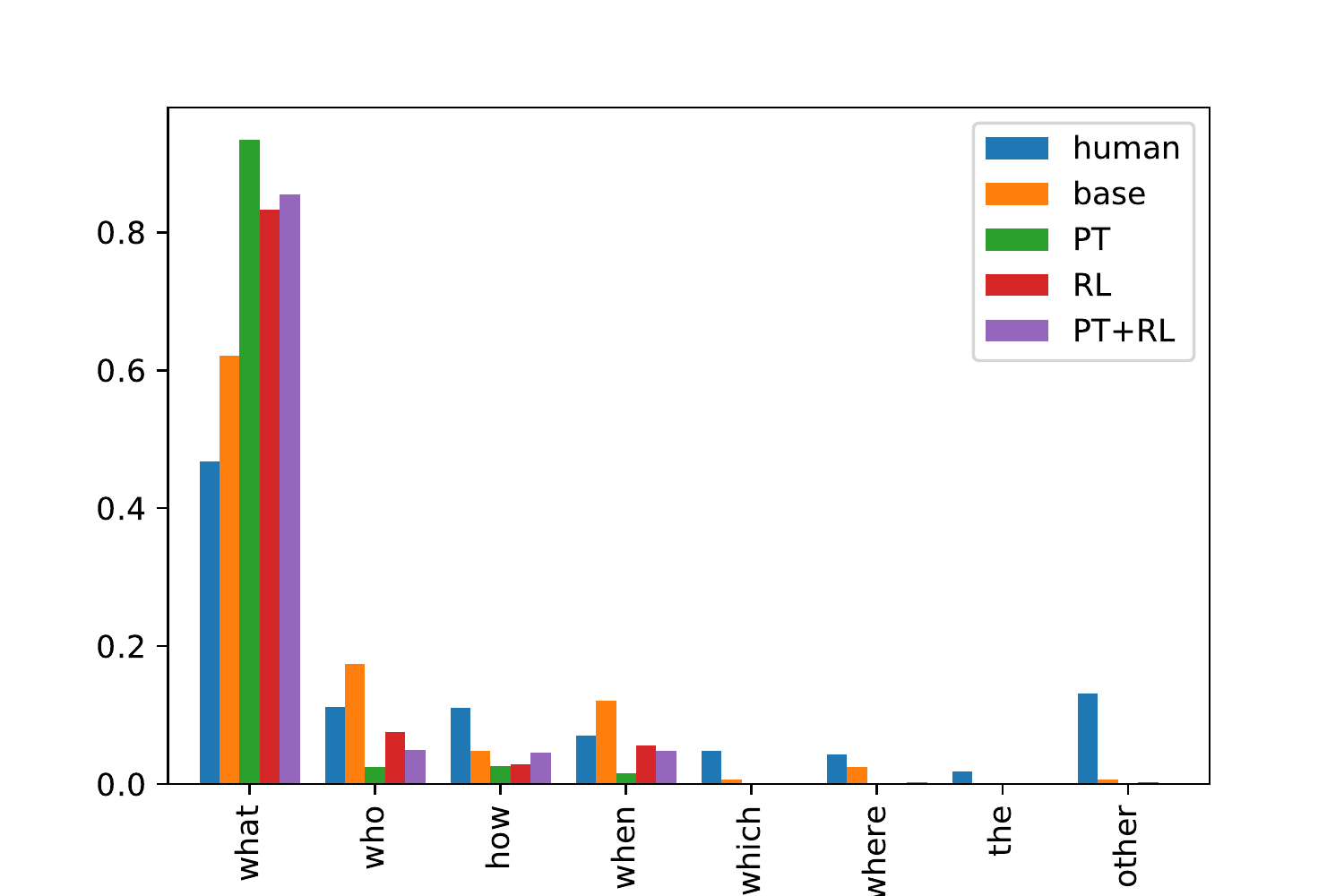}
    \includegraphics[width=\columnwidth]{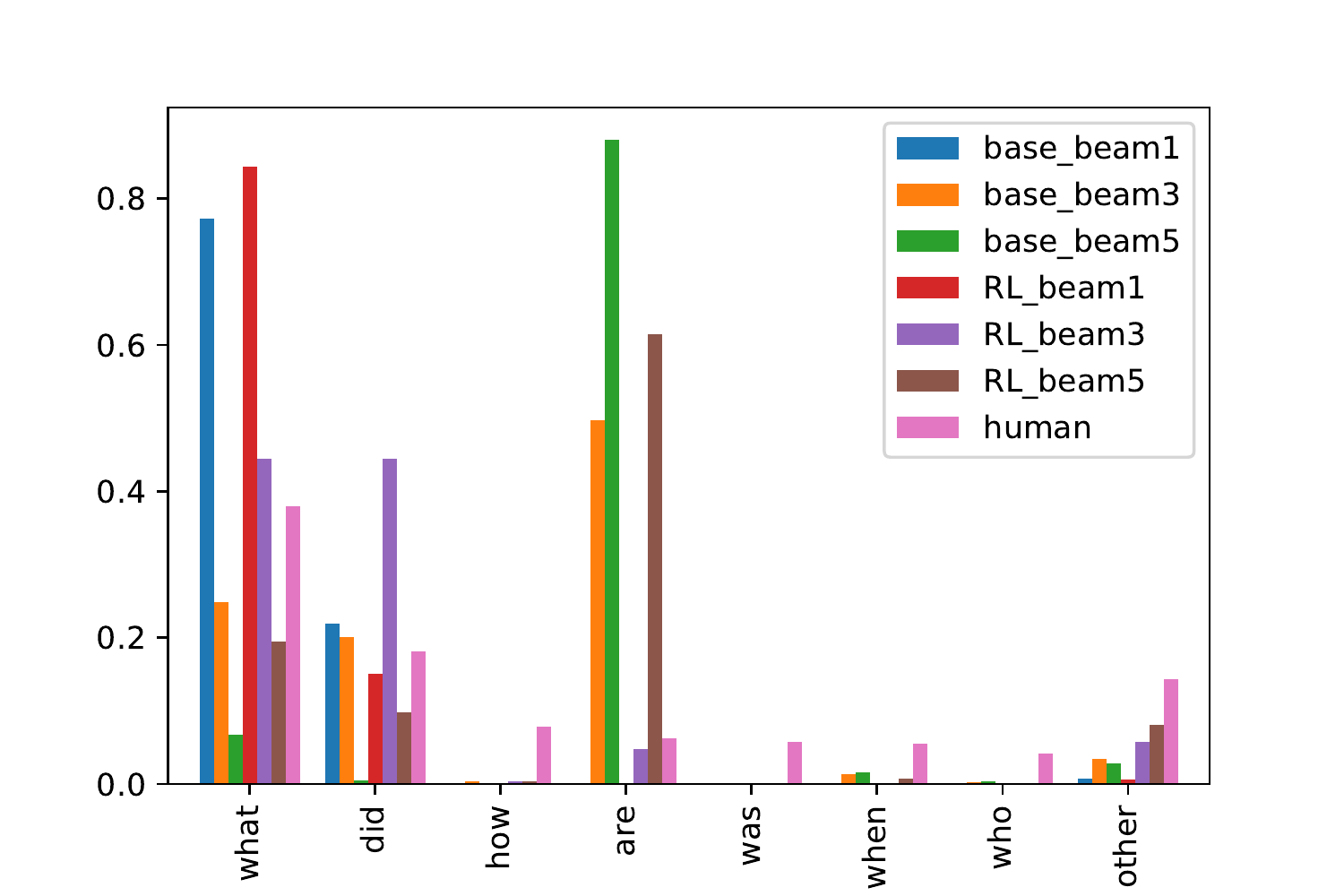}
    \caption{Distribution of the first word frequency per models for SQuAD (top) and QuAC (bottom). ``Other'' does not refer literally to the \texttt{other} token, but represents \emph{any} other token.}
    \label{fig:distribution_first_token_squad}
\end{figure*}

In Figure~\ref{fig:distribution_first_token_squad} we report the distribution of the first word frequency for the different models trained: the models without pretraining appear closer to the human-quality samples and also show more diversity. 

\subsection{Human Evaluation}
\label{subsec:human_evaluation}
\begin{table*}
\begin{center}
\begin{tabular}{|l|c|c|c|c|c|}
\hline
                     & \textbf{Answerability} & \textbf{Correctness} & \textbf{External Knowledge} & \textbf{Relevance} & \textbf{Soundness} \\ \hline \hline
\textbf{base}    & 1.23               & 4.07             & 2.41                    & 2.54           & 3.21           \\ \hline
\textbf{RL} & \textbf{1.14}      & 4.07             & \textbf{2.66}           & \textbf{2.65}  & 3.09           \\ \hline
\textbf{PT}     & \textbf{1.16}      & \textbf{4.22}    & 2.30                    & 2.43           & 3.13           \\ \hline
\textbf{PT+RL}  & 1.35               & \textbf{4.23}    & 2.21                    & 2.53           & 3.06           \\ \hline \hline
{\it human}       & {\it 1.42}               & {\it 4.61}             & {\it 2.90}                    & {\it 3.91}           & {\it 4.49}           \\ \hline
\end{tabular}
\caption{Qualitative results obtained via human evaluation.}
\label{human_evaluation}
\end{center}
\end{table*}
In addition to the automatic metrics, we proceeded to a human evaluation. We chose to use the data from our SQuAD-based experiments in order to also to measure the effectiveness of the proposed approach to derive \emph{Curiosity}-driven QG data from a standard, non-conversational, QA dataset.  
We randomly sampled 50 samples from the test set. Three professional English speakers were asked to evaluate the questions generated by:
humans (i.e. the reference questions), and models trained using pre-training (PT) or (RL), and all combinations of those methods.

Before submitting the samples for human evaluation, the questions were shuffled. Ratings were collected on a 1-to-5 likert scale, to measure to what extent the generated questions were: \emph{answerable} by looking at their context; grammatically \emph{correct}; how much \emph{external knowledge} is required to answer;
\emph{relevant} to their context; and, semantically \emph{sound}.
The results of the human evaluation are reported in Table \ref{human_evaluation}.

\section{Discussion}

\paragraph{What is the impact of the pretraining?}
We observe that for pretrained models (i.e. \emph{PT} and \emph{PT+RL}) the \emph{Correctness} is significantly higher than the models without pretraining (i.e. \emph{base} and \emph{RL}). It corroborates the higher BLEU observed for these models in Table \ref{tab:metric_results_squad}.
An other observation is that the \emph{External Knowledge} is lower for the pretrained models while the \emph{Relevance} is slightly higher. It could be due to the nature of the pretraing for which the models learn to generate non curious questions that focus on their inputs. It correlates with the significantly higher QA\_source reported in Table \ref{tab:metric_results_squad} for those pretrained models. 

\paragraph{Does Reinforcement help?}
From the human assessment we conducted -- see Table~\ref{human_evaluation}, we observe for the models trained with RL obtain higher scores for \emph{Relevance} and lower \emph{Soundness} as compared to their non-reinforced counterparts. Further, the results reported in Table~\ref{tab:metric_results_squad} show reinforced model obtaining lower BLEU and $QA_{source}$ source; conversely they score higher when it comes to $QA_{context}$. To summarize those results, we conclude that reinforcement brings improvements in terms of diversity of the generated questions, at the price of slightly degraded formulations in the outputs. 

\paragraph{How effective is our dataset creation methodology?}
Looking at the bottom row of Table~\ref{human_evaluation}, which shows the results obtained by the reference (i.e. human-generated) questions, we observe the highest relative score for all assessed dimensions, with the exception of \emph{Answerability}. This indicates that the data we derived seem to fit well the task of \emph{Curiosity}-driven question generation. As a sidenote, we remark that the models built obtain even lower scores in terms of \emph{Answerability} than humans, a fact we hypothesize due to the lower quality of the generated questions: the less \emph{sound} and \emph{correct}, the less \emph{answerable} a question would be, regardless of its context.

\begin{figure*}[!htb]
    \centering
    \includegraphics[width=.88\linewidth]{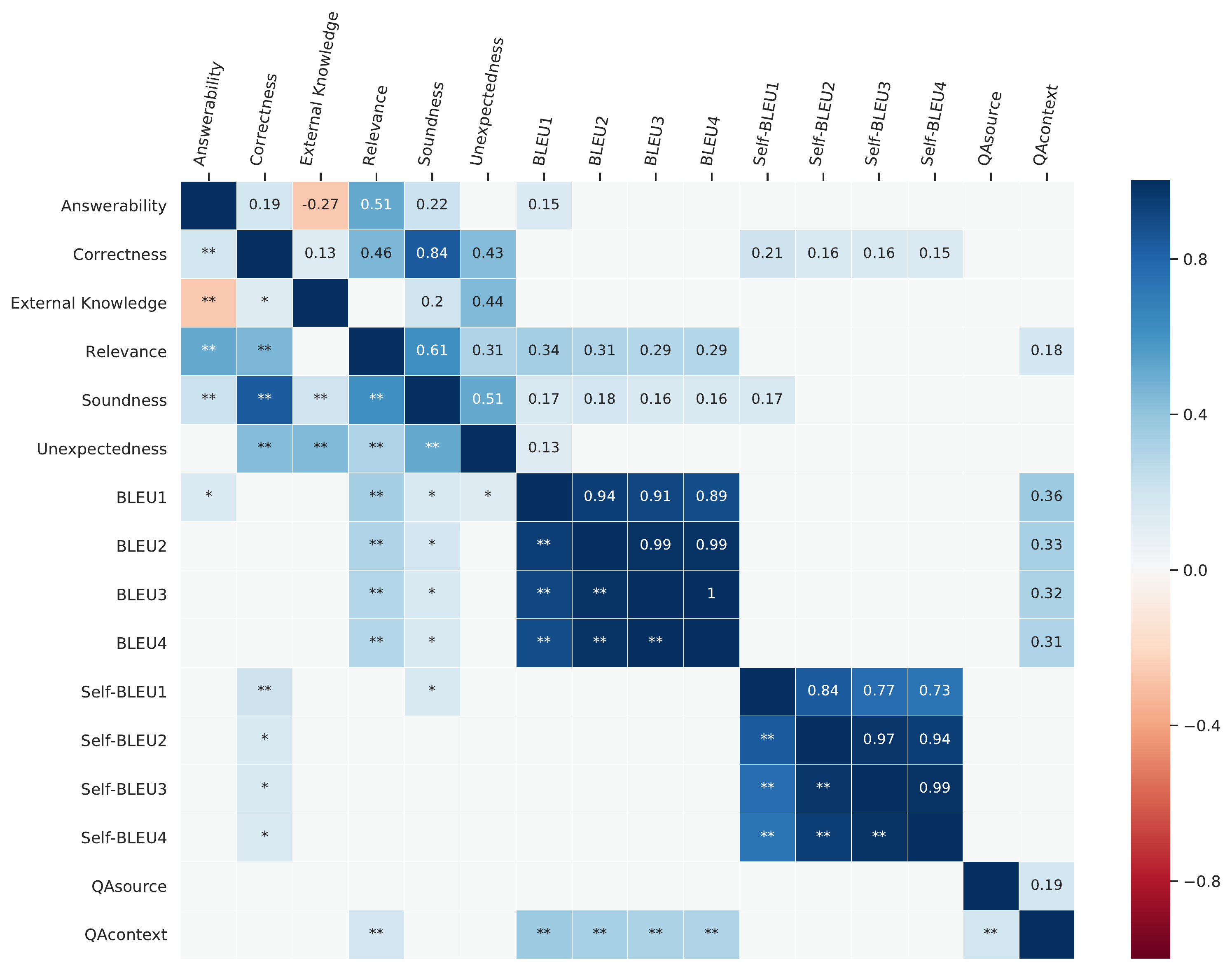}
    \caption{Correlation matrix obtained from the human assessment data ($*`: p<.05$, $**:p<.005$).}
    \label{fig:corr_mat}
\end{figure*}
\paragraph{How well do the metrics fit human judgement?}
We report the pairwise Spearman correlation and p-value among all the different metrics and human measures in Figure \ref{fig:corr_mat}.
Correlation analysis on the human assessment data shows that BLEU correlates positively with \emph{Relevance}, \emph{Answerability}, \emph{Soundness} and \emph{Unexpectedness}\footnote{To give an order of magnitude, for a standard QG task, \cite{nema_towards_2018} report a Pearson correlation of 0.258 for BLEU-1 and 0.233 for BLEU-4.}. Self-BLEU metrics correlate significantly with \emph{Soundness} and \emph{Correctness} and \emph{QA\textsubscript{context}} with \emph{Relevance}. The only human measure that does not correlate significantly with any automatic metric is \emph{External knowledge}. It is indeed one of the most challenging aspect to evaluate, even for humans. However, as expected, it correlates negatively with \emph{Answerability}. 

\section{Conclusions}
The human skill of asking inquisitive questions allows them to learn from the other and increase their knowledge. \emph{Curiosity}-driven question generation could be a key component for several human-machine interaction scenarios. We thus proposed a new task: \emph{Curiosity}-driven Question Generation. In absence of data directly usable for this task, we propose an automatic method to derive it from conversational QA datasets. Recognizing that the great majority of QA datasets are not dialogue-based, we also extend the method to standard QA data. Our experiments, including strategies as pretraining and reinforcement, show promising results under both automatic and human evaluation.

In future works, we plan to extend the approach to conditional generation of \emph{Curiosity}-driven questions. 

\clearpage
\appendix
\onecolumn

\section{Computational Costs}
All our experiments were run on a single nVidia 2080ti gpu. For SQuAD experiments, training time amounted to circa 45 minutes and 12 hours for the model built without and with reinforcement, respectively. The additional pretraining step took roughly 2 hours. For QuAC experiments, training time amounted to circa 2 hours and 15 hours for the models built without and with reinforcement, respectively.

\section{Sample Outputs}
\paragraph{From QuAC (test set):}

\begin{center}
\noindent\fbox{%
    \parbox{.96\columnwidth}{
        \textbf{Context ($P'$)}:\\
        Discovery in the United Kingdom\\
        The Seekers were offered a twelve-month position as on-board entertainment on the Sitmar Line passenger cruise ship Fairsky in March 1964. In May, they travelled to the U.K. and had intended to return to Australia after staying ten weeks, but upon arrival they were offered work by a London booking agency, the Grade Organisation.\\
        
        \textbf{Model $\Rightarrow$ Outputs:}\\
        base\_beam1  $\Rightarrow$  what was the name of the band ?\\
        base\_beam3  $\Rightarrow$  are there any other interesting aspects about this article ?\\
        base\_beam5  $\Rightarrow$  are there any other interesting aspects about this article ?\\
        RL\_beam1  $\Rightarrow$  what was the name of the album ?\\
        RL\_beam3  $\Rightarrow$  did they have any other albums ?\\
        RL\_beam5  $\Rightarrow$  are there any other interesting aspects about this article ?\\
        
        \textbf{Human reference:}\\
        human  $\Rightarrow$ what else can you tell me about thier discovery ?

    }%
}

\noindent\fbox{%
    \parbox{.96\columnwidth}{
        \textbf{Context ($P'$)}:\\
        1977-1980: Death of a Ladies' Man and End of the Century\\
        Phillip Harvey Spector (born Harvey Phillip Spector, December 26, 1939) is an American record producer, musician, and songwriter who developed the Wall of Sound, a music production formula he described as a "Wagnerian" approach to rock and roll. Spector is considered the first auteur among musical artists for the unprecedented freedom and control he had over every phase of the recording process. Additionally, he helped engender the idea of the studio as its own distinct instrument. For these contributions, he is acknowledged as one of the most influential figures in pop music history.

        \textbf{Model $\Rightarrow$ Outputs:}\\
        base\_beam1  $\Rightarrow$  what was his first album ?\\
        base\_beam3  $\Rightarrow$  what happened in 1985 ?\\
        base\_beam5  $\Rightarrow$  are there any other interesting aspects about this article ?\\
        RL\_beam1  $\Rightarrow$  what was the name of the album ?\\
        RL\_beam3  $\Rightarrow$  what was the name of the album ?\\
        RL\_beam5  $\Rightarrow$  did he have any other albums ?\\
        
        \textbf{Human reference:}\\
        human  $\Rightarrow$  was death of a ladies man an album ?

    }%
}
\end{center}
\newpage
\paragraph{From SQuAD (test set):}

\begin{center}
\noindent\fbox{%
    \parbox{.96\columnwidth}{
        \textbf{Context ($P'$)}:\\
        The Broncos defeated the Pittsburgh Steelers in the divisional round, 23–16, by scoring 11 points in the final three minutes of the game.\\
        
        \textbf{Model $\Rightarrow$ Outputs:}\\
        base  $\Rightarrow$  who was the head of the steelers ?\\
        PT  $\Rightarrow$  what was the name of the game ?\\
        RT  $\Rightarrow$  when was the broncos game ?\\
        PT+RT  $\Rightarrow$  what was the name of the steelers ?\\
        
        \textbf{Human reference:}\\
        human  $\Rightarrow$ how many seconds were left in the game when the broncos intercepted the pass that won the game ?
    }%
}

\noindent\fbox{%
    \parbox{.96\columnwidth}{
        \textbf{Context ($P'$)}:\\
        More than 1 million people are expected to attend the festivities in San Francisco during Super Bowl Week.\\

        \textbf{Model $\Rightarrow$ Outputs:}\\
        base  $\Rightarrow$  how many people live in san diego ?\\
        PT  $\Rightarrow$  how many people live in san diego ?\\
        RT  $\Rightarrow$  what is the average rainfall in san diego ?\\
        PT+RT  $\Rightarrow$  how many people live in san diego ?\\
        
        \textbf{Human reference:}\\
        human  $\Rightarrow$  who is the mayor of san francisco ?

    }%
}
\end{center}

\clearpage
\twocolumn
\bibliographystyle{acl_natbib}
\bibliography{reciTAL}
\end{document}